\documentclass[acmsmall,screen, nonacm]{acmart}
\AtBeginDocument{%
  }

\usepackage{multirow}
\usepackage{caption}
\usepackage{subcaption}

\DeclareMathOperator{\EX}{\mathbb{E}}

\begin{document}

\title{Non-Uniform Quantisation for 3DGS Compression}

\author{Bert Van hauwermeiren}
\email{bert.karel.van.hauwermeiren@vub.be}
\orcid{0009-0002-6707-3434}
\affiliation{%
  \institution{Vrije Universiteit Brussel}
  \city{Brussels}
  \country{Belgium}
}

\author{Patrice Rondao Alface}
\email{patrice.rondao_alface@nokia.com}
\orcid{0000-0002-5402-6180}
\affiliation{%
  \institution{Nokia}
  \city{Antwerp}
  \country{Belgium}
}

\author{Adrian Munteanu}
\email{adrian.munteanu@vub.be}
\orcid{0000-0001-7290-0428}
\affiliation{%
  \institution{Vrije Universiteit Brussel}
  \city{Brussels}
  \country{Belgium}
}

\begin{abstract}
3D Gaussian Splatting (3DGS) has emerged as a powerful technique for novel view synthesis, yet its high bitrate requirements pose significant challenges for storage and transmission. To enable practical applications and ensure interoperability within the 3DGS ecosystem, standardised compression formats are essential. In this paper, we propose a novel non-uniform quantisation scheme specifically tailored for 3DGS models. Our approach adapts to the underlying data distribution by applying importance-weighted quantisation and eliminating post-voxelisation redundancy through importance weighted merging. Extensive evaluations on benchmark datasets demonstrate that our method achieves state-of-the-art compression performance. Furthermore, the proposed scheme is compatible with any point-cloud-based representation and is intended as a formal contribution to the upcoming MPEG 3DGS compression standardisation activities.
\end{abstract}



\keywords{3DGS, compression, quantisation, voxelisation}


\settopmatter{printacmref=false} 
\setcopyright{none}              
\renewcommand\footnotetextcopyrightpermission[1]{} 

\maketitle

\section{Introduction}

3D Gaussian Splatting (3DGS)~\cite{kerbl_3d_2023} has recently emerged as the state-of-the-art paradigm for novel view synthesis. By representing scenes as collections of anisotropic 3D Gaussians, 3DGS enables real-time rendering with visual fidelity that often surpasses coordinate-based radiance fields. Each Gaussian is defined by five key attributes: 3D position, scale, rotation, opacity, and spherical harmonic (SH) coefficients. However, this explicit representation comes at a significant cost; up to millions of Gaussians can be required to represent a single scene, resulting in massive bitrates that hinder real-world deployment in streaming, virtual reality, and mobile environments.

While recent research has focused on developing 3DGS training methods with a reduced memory footprint, through minimising the number of Gaussians and representing them more efficiently (e.g., using anchor-based generation~\cite{lu_scaffold-gs_2024} or hash-grid contexts~\cite{chen_hac_2025}), the industry requires standardised compression formats to enable interoperability across tools, platforms, and vendors. This closely mirrors earlier successful developments in multimedia, where standardised image, audio, and video compression formats enabled capturing content on virtually any device and seamlessly opening, editing, and distributing it on any other device, resulting in a smooth and consistent user experience. Such standardisation lowered adoption barriers, reduced ecosystem fragmentation, and provided long-term stability for content and tooling.

In response, the Motion Pictures Expert Group (MPEG) has initiated two fast-track standardisation activities by adapting the Video-based Point Cloud Compression (V-PCC)~\cite{isoiec_isoiec_2025, alface_coding_2025} and Geometry-based Point Cloud Compression (G-PCC)~\cite{isoiec_isoiec_2023} standards for 3DGS, leveraging the structural similarities between point clouds and 3D Gaussian representations. The G-PCC standard utilises an octree or predictive-tree structure to encode the voxelized 3DGS geometry, and a Region Adaptive Hierarchical Transform (RAHT) or lifting transform, followed by entropy coding for the quantised attributes. The V-PCC standard, on the other hand, divides the Gaussians into patches, which are subsequently mapped onto 2D frames that are encoded using existing highly efficient video codecs, such as HEVC~\cite{sullivan_overview_2012} or VVC~\cite{bross_overview_2021}. Both codecs apply uniform quantisation to the 3DGS geometry and attributes before entropy coding. This approach is suboptimal, as it fails to account for the non-uniform distribution of Gaussian data and the varying perceptual importance of individual Gaussians.

To address these limitations, we propose a novel non-uniform 3DGS quantisation pipeline, built upon earlier research in point cloud voxelisation~\cite{van_hauwermeiren_non-uniform_2025}. We are the first to develop a non-uniform, importance-weighted quantisation framework for both geometry and attributes that is fully compatible with standardised point-cloud codecs. Our primary contributions are:
\begin{itemize}
    \item A novel non-uniform quantisation scheme designed to minimise the weighted reconstruction error. By integrating an importance metric, we prioritise high-impact Gaussians to optimise the rate-distortion tradeoff.
    \item A comparison of several novel importance-weighted merging approaches for Gaussians that become co-located after voxelisation.
    \item An extensive performance analysis conducted in accordance with MPEG's 3DGS common test conditions (CTC). The results demonstrate that the proposed method consistently outperforms current state-of-the-art methods, providing significant gains in the rate-distortion trade-off across diverse scene types. The weighted merging contribution has already been adopted in the V-PCC Amd1 for GS.
\end{itemize}

While our method is developed to be a contribution to the upcoming MPEG 3DGS standards, the underlying logic remains compatible with any point-cloud-based splatting representation and entropy-coding backend. The rest of the paper is structured as follows: we situate the proposed method in relation to other related works in Section~\ref{sec:related}. We then provide a detailed explanation of the proposed method's design in Section~\ref{sec:proposed}, followed by an enumeration of all the experiments done in Sections~\ref{sec:experiments}~and~\ref{sec:abl}. Finally, a discussion and conclusion of the research are given in Sections~\ref{sec:discussion}~and~\ref{sec:conclusion}.

\section{Related Works}
\label{sec:related}

3D Gaussian Splatting (3DGS)~\cite{kerbl_3d_2023} has redefined the landscape of novel view synthesis by utilising a set of anisotropic 3D Gaussians for scene representation. Unlike the implicit coordinate-based mapping used in Neural Radiance Fields (NeRF) \cite{mildenhall_nerf_2021}, 3DGS offers an explicit representation that enables high-fidelity rendering at real-time frame rates. Each Gaussian primitive $g$ is parameterised by its mean or position $\mu_g$, rotation $q_g$, scale $s_g$, opacity $o_g$ and spherical harmonic coefficients $c_g$ that encode view-dependent appearance. The final rendered colour $C(p)$ of each pixel $p$ is computed via alpha-blending of $N$ ordered Gaussians overlapping that pixel:

\begin{equation}
\label{eq:render}
    C(p) = \sum_{g_i} C_{g_i} \alpha_{g_i} \prod_{j=1}^{i-1} (1-\alpha_{g_j}) = \sum_g w_{p,g} C_g,
\end{equation}
where $\alpha_g$ is the density of the Gaussian at that specific pixel and $w_{p,g}$ is the per-Gaussian contribution that pixel. The rendered images are compared against ground-truth training views, and all Gaussian parameters are optimised via gradient descent. When combined with iterative densification and pruning, this framework achieves state-of-the-art rendering quality at real-time rendering speeds. However, the requirement for up to millions of primitives per scene results in a prohibitive memory footprint, necessitating advanced compression strategies.

\paragraph{Compressed representations.} A first line of research seeks to retain the core 3DGS rendering pipeline while significantly reducing its memory footprint through altering the representation or training loop, resulting in smaller representations. Common strategies include reducing the number of Gaussians~\cite{lee_compact_2024, fang_mini-splatting_2024}, lowering the SH order~\cite{zhu_low-rank_2026}, exploiting spatial and parametric redundancy, and applying quantisation. EAGLES~\cite{girish_eagles_2025} introduces vector quantisation for Gaussian attributes and prunes primitives based on an influence metric that estimates their contribution to rendered views. LFGS~\cite{wu_lfgs_2025} combines vector quantisation with pruning criteria based on Gaussian volume and opacity, while dynamically adapting the SH degree per primitive to balance fidelity and storage. CompGS~\cite{navaneet_compgs_2025} further explores redundancy by applying K-means-based vector quantisation to Gaussian parameters during training and encoding the resulting cluster indices with run-length-style compression. A significant shift toward hierarchical structures is seen in Scaffold-GS~\cite{lu_scaffold-gs_2024}, which replaces explicit per-Gaussian storage with a sparse set of spatial anchors whose learned features are decoded by a neural network into local Gaussian clusters. HAC~\cite{chen_hac_2025} extends this by utilising hash-grid interpolation for anchor features and implementing adaptive, per-anchor quantisation to allocate bits where they are most perceptually relevant. While achieving impressive rate-distortion performance, these methods require modified representations and renderers, which are not allowed in the current G-PCC and V-PCC amendments.

\paragraph{Alternative representations.} Beyond compressing standard volumetric Gaussians, several works propose alternative splatting primitives that are more expressive or compact. 2D Gaussian Splatting~\cite{huang_2d_2024} replaces volumetric Gaussians with planar disks, enabling view-consistent geometry modelling that substantially reduces the number of required primitives and inspiring further planar primitive research~\cite{held_triangle_2025, zhang_quadratic_2025, jurca_fourier_2026}. Deformable Beta Splatting~\cite{liu_deformable_2025} generalises the Gaussian kernel to a deformable beta distribution, allowing primitives to become "flatter" and better aligned with scene geometry; applying the same kernel to colour enables improved modelling of reflections and specularities, further reducing the total primitive count. GES~\cite{hamdi_ges_2024} employs Generalised Exponential Functions to represent scene content. These functions can capture sharp edges and high-frequency details that the low-pass nature of traditional Gaussians often blurs, thereby enabling a drastic reduction in bitrate. However, these alternative primitives don't bring bitrate reduction on par with more advanced compression techniques.

\paragraph{Post-training compression.} A final class of methods focuses on post-training compression of already optimised 3DGS models. Some approaches directly apply established point cloud compression standards, such as V-PCC and G-PCC, to Gaussian parameters~\cite{alface_coding_2025, bang_mpeg_2026}. FlexGaussian~\cite{tian_flexgaussian_2025} introduces attribute-aware compression by pruning and quantising attributes independently, prioritising geometrically critical parameters such as position, rotation, and scale over higher-order SH coefficients, and allocating different bit-widths per attribute channel. Meson-GS~\cite{xie_mesongs_2025} similarly prunes Gaussians using an importance metric derived from volume, opacity, and rendering contribution. In addition to classical point cloud coding techniques, positions are encoded using an octree, while important attributes are transformed using region-adaptive hierarchical transforms (RAHT)~\cite{de_queiroz_compression_2016}. Higher-order SH coefficients are vector-quantised, and rotations are represented using Euler angles. To enhance the rate-distortion efficiency of uniform voxelisation, adaptive voxelization~\cite{wang_adaptive_2025} introduces a variable depth voxelisation strategy. This method employs a subdivision heuristic that evaluates the local density of Gaussians, their respective volumes, and their distance to the voxel center to determine the optimal level of refinement. These methods demonstrate that substantial reductions in storage and bandwidth are possible without modifying the original 3DGS training process.

\section{Proposed Method}
\label{sec:proposed}

\begin{figure}[t]
    \centering
    \includegraphics[width=1.0\linewidth]{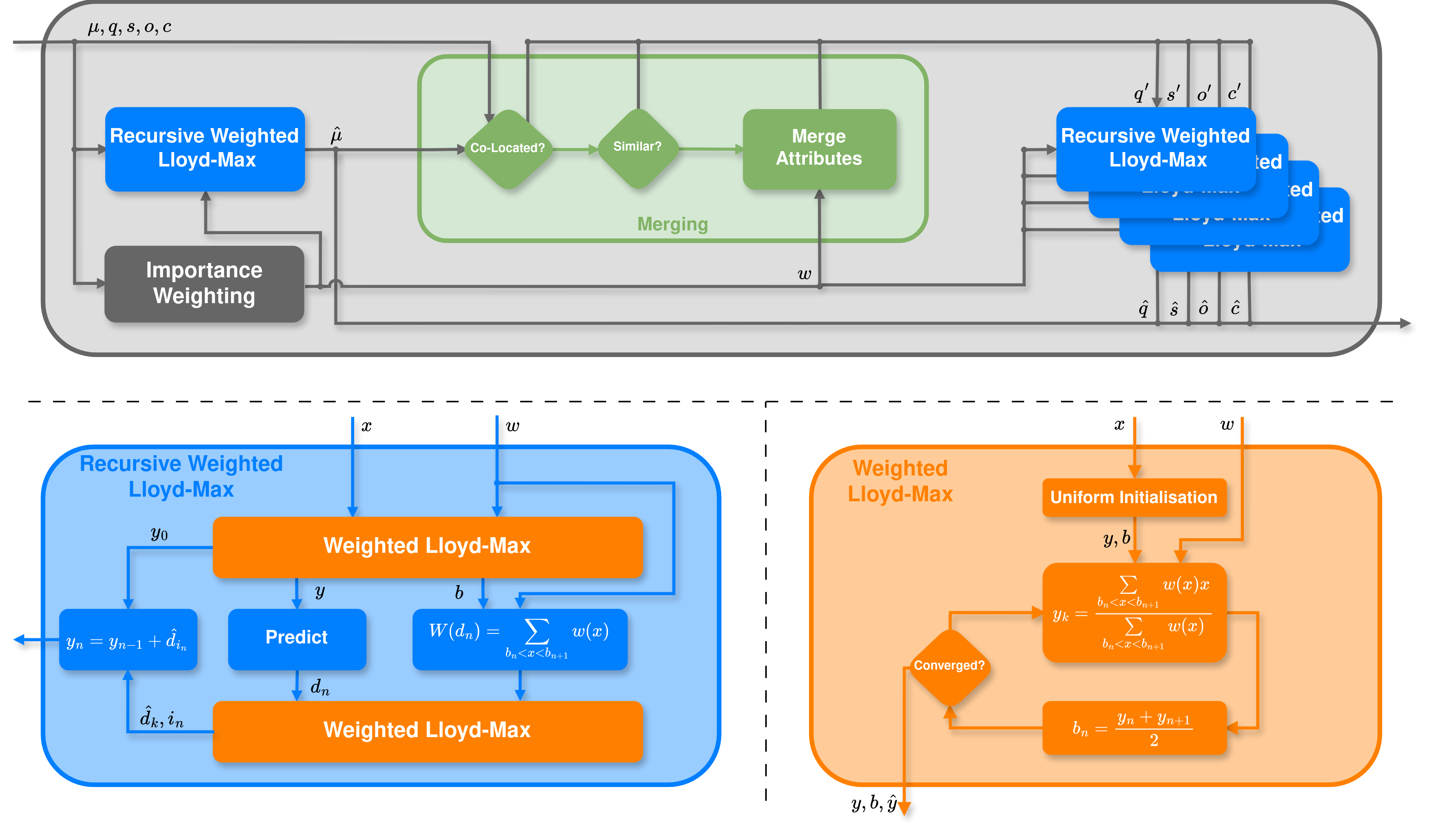}
    \caption{Overview of the proposed 3DGS quantisation pipeline. The full workflow begins \textbf{(Top)} with high-speed importance estimation for each primitive. This metric directly guides the subsequent non-uniform quantisation of the Gaussian positions (means). Primitives that are co-located within the resulting voxel grid are collapsed via an importance-weighted geometric merging module, combining all attributes into a single representative Gaussian. The remaining attributes are then also processed by our non-uniform quantiser \textbf{(Bottom)}, which adapts the Lloyd-Max algorithm \textbf{(Bottom-Right)} to minimise importance-weighted Mean Squared Error (WMSE) and recursively compresses the resulting reconstruction values \textbf{(Bottom-Left)} to achieve substantial metadata bitrate savings.}
    \label{fig:pipeline}
\end{figure}

In this section, we present our proposed non-uniform quantisation pipeline designed for 3DGS. In Subsection~\ref{subseq:importance_weighting}, we discuss our importance weighting strategy; this metric allows the pipeline to minimise the distortion made after rasterisation rather than the isolated per-Gaussian parameter error. Building on these weights, Subsection~\ref{subsec:wrlm} details the core of our framework: a recursive weighted Lloyd-Max module designed for the non-uniform quantisation of both 3DGS positions and attributes. Finally, Subsection~\ref{subsec:merging} explores various strategies for the weighted merging of co-located Gaussians—those sharing the same voxel—to effectively reduce scene redundancy while preserving rendering quality. The complete architecture and logical flow of this pipeline are visualised in Figure~\ref{fig:pipeline}.

\subsection{Importance Weighting}
\label{subseq:importance_weighting}

The proposed non-uniform quantisation method minimises distortion for each Gaussian subject to a weighting function. To optimise for post-rendering fidelity, this weighting must accurately reflect the influence of individual Gaussians on the final rendered images. Using the rendering equation~\ref{eq:render}, we can express the rendered image quantisation error $D$ as a function of the pixel-wise per-Gaussian contributions $w_{p,g}$ (Equation~\ref{eq:render}) and the per-Gaussian quantisation error $\epsilon_g = C_g - \hat{C}_g$:
\begin{align*}
D &= \frac{1}{N} \sum_p \left( C(p) - \hat{C}(p)\right)^2 \\
        &= \frac{1}{N} \sum_p \left( \sum_g w_{p, g} \epsilon_g \right)^2 \\
        &= \frac{1}{N} \sum_p \sum_g w_{p, g}^2 \epsilon_g^2 + \frac{1}{N} \sum_p \sum_{g \neq g'} w_{p,g} w_{p,g'} \epsilon_g \epsilon_{g'}.
\end{align*}
To derive a practical per-Gaussian error metric, we analyse the expected value of the total distortion $\EX[D]$. Directly evaluating the cross-terms is generally intractable because the quantisation errors $\epsilon_g$ and $\epsilon_{g'}$ are deterministic functions of the input signals. If two Gaussians are spatially or statistically correlated, their quantisation errors are not guaranteed to be independent. Assuming that the quantisation steps are small relative to the signal variation and that the source density function is locally smooth, we can invoke high-rate quantisation theory~\cite{gersho_vector_1992} to model the quantisation errors $\epsilon_g$ as independent, zero-mean random variables. Taking the expectation of the distortion yields:\begin{align*}
    \EX[D] &= \frac{1}{N} \sum_p \sum_g w_{p, g}^2 \EX[\epsilon_g^2] + \frac{1}{N} \sum_p \sum_{g \neq g'} w_{p,g} w_{p,g'} \EX[\epsilon_g \epsilon_{g'}].
\end{align*}
Because the errors are independent and zero-mean under these conditions, their joint expectation vanishes:
\begin{align*}
    \EX[\epsilon_g \epsilon_{g'}] &= \EX[\epsilon_g] \EX[\epsilon_{g'}] = 0 \quad \text{for } g \neq g'.
\end{align*}
Consequently, the cross-terms can be neglected, simplifying the expected total reconstruction error to:
\begin{align*}
    \EX[D] \approx \frac{1}{N} \sum_g \left(  \sum_p w_{p, g}^2 \right) \EX[\epsilon_g^2].
\end{align*}
In conclusion, we find that the error for each Gaussian contributes to the final rendering with a factor 
\begin{equation}
\label{eq:importance}
    w_g = \sum_p w_{p,g}^2 = \sum_p \alpha_i^2 \prod_{j=1}^{i-1}(1-\alpha_j)^2.
\end{equation}
This per-Gaussian importance metric $w_g$ can be easily derived from the per-pixel contributions $w_{p,g}$, which are already computed during pixel-wise rasterisation. However, because full-scene rasterisation is computationally expensive, we propose a high-speed alternative: an approximate weighting function that predicts importance based on each Gaussian's scale, opacity and position. Given the complex, non-linear relationship between a Gaussian’s geometric parameters and its final visibility, we employ a tiny, generalisable Multi-Layer Perceptron (MLP) to estimate this value:
\begin{equation}
    \hat{w}_g = \text{MLP}\left( \prod_{i=1}^{3} s_{g,i}, o_g, \mu_g \right).
\end{equation}
On the choice of inputs: Equation~\ref{eq:importance} shows that spherical harmonics are not used. We assume that viewpoints are uniformly distributed across the scene, so the rotation of the geometry will not affect the importance, from testing this assumption also works for scenes that are primarly seen from one angle. Lastly, Gaussians that are closer to the center of the scene are more likely to be part of the foreground, thus appearing in more viewpoints and likely being larger in each. These measures of Gaussian importance are subsequently used by each of the following quantisation steps.

\subsection{Weighted Recursive Lloyd-Max Quantisation}
\label{subsec:wrlm}

The central component of the proposed pipeline is a non-uniform quantisation module designed to efficiently compress 3DGS geometry and attributes by accounting for the specific statistical distribution and perceptual importance of each Gaussian. While the classical Lloyd-Max algorithm typically minimises the standard Mean Squared Error (MSE), our approach optimises a weighted Mean Squared Error (WMSE):
\begin{align*}
    D = \EX[w(x)(x-Q(x))^2] = \sum_{i=0}^{n-1} \sum_{x \in  [b_i, b_{i+1})} w(x)(x-y_i)^2,
\end{align*}
where $y$ and $b$ are the quantiser reconstruction values and boundaries, respectively, while $w(x)$ is the importance of the Gaussian associated with the scalar value $x$. To minimise this objective, the reconstruction values at each iteration are updated to the importance-weighted centroid of their respective bins:
\begin{align*}
y_i = \frac{\sum_{x \in  [b_i, b_{i+1})} w(x) x}
            {\sum_{x \in  [b_i, b_{i+1})} w(x)}
\end{align*}
This is interesting because the goal of 3DGS compression is not to reconstruct the 3D Gaussians as closely as possible, but rather to maintain the quality of the rasterised images. Some Gaussians have a bigger impact on the image generation than others; These Gaussians should thus also have a larger weight in the quantisation process. This weighted non-uniform quantiser allows for significantly higher image reconstruction quality; however, the bitrate overhead of the quantiser metadata limits the rate-distortion performance. For a given bit depth $b$, we need to communicate $2^b$ floating-point reconstruction values to dequantise at the decoder and to achieve reconstructions of acceptable quality, high bit depths are required. To significantly reduce this meta-data overhead, we quantise the differences $d_i=y_i-y_{i-1}$, yielding an approximation of the initial weighted Lloyd-Max quantiser. The goal is to minimise the additional distortion $\Hat{D}$:
\begin{align*}
\Hat{D} &= \EX[w(x)(x-Q(x))^2] - \EX[w(x)(x-\Hat{Q}(x))^2] \\
        &= \cdots = \sum_{i=0}^{n-1} \sum_{x \in  [b_i, b_{i+1})}  w(x)(y_i-\hat{y}_i)^2 \\
        &= \sum_{k=0}^{K-1} \sum_{d_i \in  [b'_k, b'_{k+1})} \sum_{x \in  [b_i, b_{i+1})} w(x) (d_i - \hat{d}_k)^2 \\
        &= \sum_{k=0}^{K-1} \sum_{d_i \in  [b'_k, b'_{k+1})} W(d_i) (d_i - \hat{d}_k)^2
\end{align*}
Note that distortion-minimising second quantisation comes down to another weighted Lloyd-Max iteration using a different number of quantisation bits, and the importance weight for each difference $d$ is the sum of the importance of all the values in the given bin $W(d_i) = \sum_{x \in  [b_i, b_{i+1})} w(x)$. The metadata that needs to be communicated to the decoder are the first floating point reconstruction value $y_0$, the floating point difference reconstruction values $\{\hat{d}_k | k=0..K-1\}$ and the $log_2(K)$-bit difference assignments $\{k_i | i=0..n-1\}$. The decoder can then calculate the reconstruction values as 
\begin{align*}
 y_i = \begin{cases}
            y_0, & i=0 \\
			y_{i-1} + \hat{d}_{k_i}, & i > 0. \\
		 \end{cases}
\end{align*}
The overhead is reduced from $32n$ to $32(K+1) + 
log_2(K)n$, which for typical values of $K=4$ and $n$ between $2^{10}$ and $2^{18}$, results in large bitrate savings, at a cost of small reductions of reconstruction quality and increase in computational complexity.

\subsection{Weighted Merging}
\label{subsec:merging}

After voxelisation, i.e. the quantisation of 3DGS positions, a significant number of Gaussians share the same quantised mean with other Gaussians. Consolidating these primitives via merging significantly reduces the number of primitives and thus the total bitrate, while incurring only a marginal loss in reconstruction quality. However, since the union of multiple ellipsoids does not yield a standard ellipsoid, the merging process results in an approximation of the original Gaussians. Rather than relying on simple attribute averaging followed by costly fine-tuning iterations \cite{wang_adaptive_2025}, we propose a series of importance-weighted merging strategies designed for high accuracy without retraining. We evaluate three distinct methods for geometric merging, which are visualised in Figure~\ref{fig:interp}:

\begin{figure}[t]
    \centering
    \includegraphics[width=0.6\linewidth]{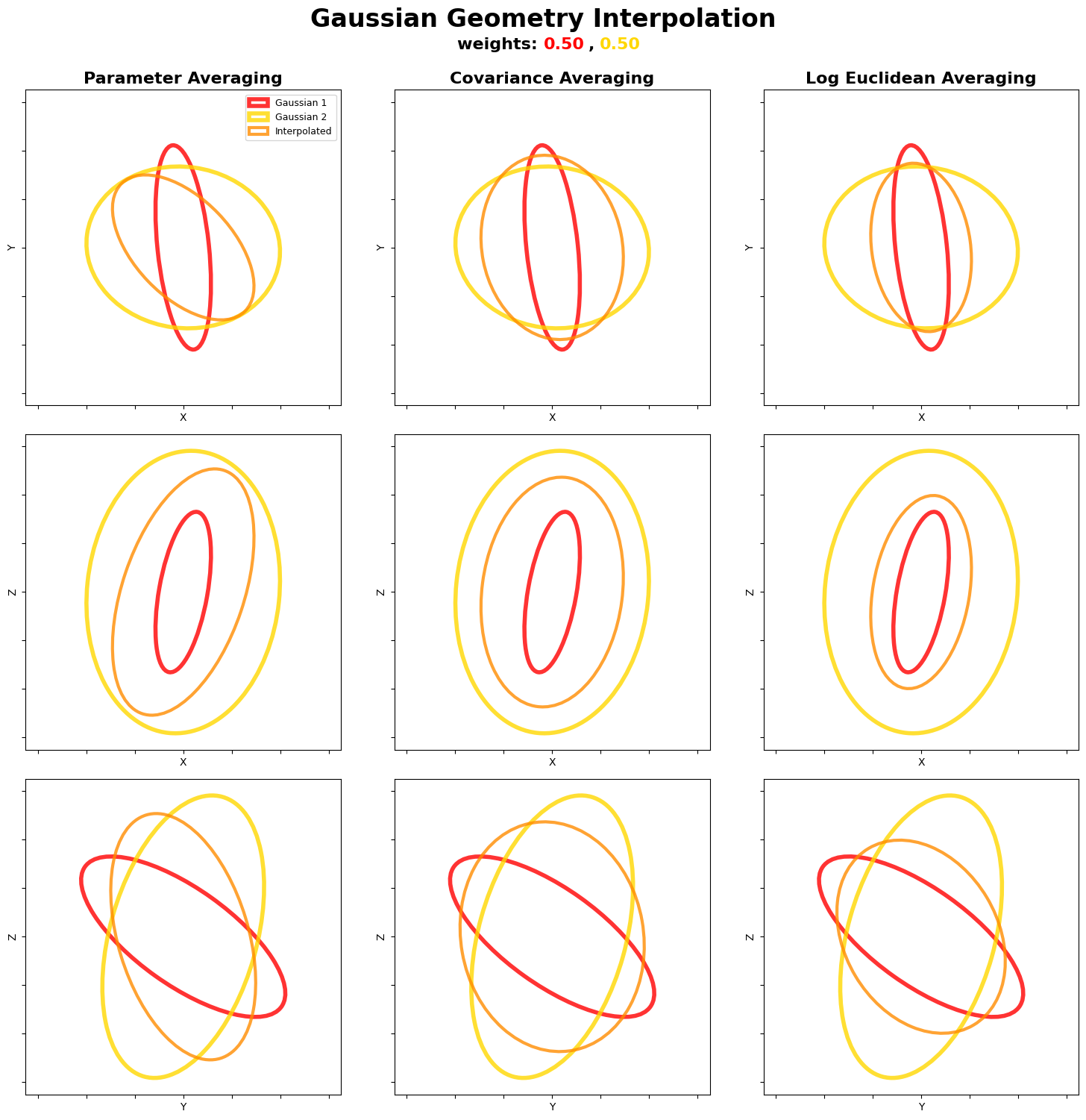}
    \caption{Visualisation of the different merging strategies for equally important source Gaussians.}
    \label{fig:interp}
\end{figure}

\begin{enumerate}

    \item \textbf{Weighted Parameter Averaging} The first approach computes the weighted mean of scales and rotations directly in their parameter space. A straightforward linear interpolation of quaternions, such as:
    \begin{equation}
        \overline{q}_{\text{linear}} = \frac{1}{\sum_{i=1}^n w_i} \sum_{i=1}^n w_i q_i
    \end{equation}
    fails to respect the antipodal symmetry of unit quaternions, where $q$ and $-q$ represent identical spatial rotations. Consequently, arbitrary sign assignments can lead to destructive interference and severe wrap-around artifacts. To account for this double-cover property and properly eliminate bias, we adopt the optimization-based averaging scheme proposed by Markley~et~al.~\cite{markley_averaging_2007}:
    \begin{equation} 
        \overline{q} = \arg\max_{q} q^T \left( \sum_{i=1}^n w_i q_i q_i^T \right) q, 
    \end{equation}
    where $q_i \in \mathbb{R}^4$ denotes the column vector representation of the $i$-th quaternion, and $w_i$ is its corresponding importance weight.
    
    \item \textbf{Weighted Covariance Space Averaging} The second strategy involves a weighted average performed in the covariance space:
    \begin{equation} 
    \overline{\Sigma} = \frac{1}{ \sum_{i=1}^n w_i} \sum_{i=1}^n \boldsymbol{\Sigma_i} w_i, 
    \end{equation} 
    where $\Sigma = RSS^TR^T$. We then apply eigenvalue decomposition to extract the merged scales (eigenvalues) and rotation (eigenvector matrix) from $\Sigma$. While eigenvalue decomposition does not yield a unique solution, any solution is a valid representation; we enforce consistency to reduce entropy. Although applying this transformation and decomposition to non-merged Gaussians could potentially improve entropy coding efficiency by normalising the representation, it comes with significant computational overhead and is out of scope of this work.

    \item \textbf{Weighted Log-Euclidean Averaging} To better preserve the geometric structure of the Symmetric Positive Definite matrices, we consider the weighted average under the affine-invariant Riemannian metric. Since a closed-form solution is not always available for this metric, we utilise the Log-Euclidean mean as a robust approximation~\cite{arsigny_geometric_2007}: 
    \begin{equation} 
    \overline{\Sigma} = \exp\left( \frac{1}{\sum_{i=1}^n w_i} \sum_{i=1}^n \log(\boldsymbol{\Sigma_i})w_i \right). 
    \end{equation}
\end{enumerate}

Lastly, the opacity and spherical harmonics of the merged primitive are computed as the weighted average of their respective input values. We observe that merging Gaussians that are both highly influential and visually distinct can cause severe rendering degradation. To mitigate this, we introduce a dissimilarity metric and restrict the merging process to pairs falling below a predefined threshold:
\begin{equation}
    \text{dissimilarity}_{i,j} = \lambda_1 \left\Vert SH_i - SH_j \right\Vert + \lambda_2 \left\Vert \Sigma_i - \Sigma_j \right\Vert + \lambda_3 \left|o_i - o_j\right| + \lambda_4 \min\left(w_i, w_j\right),
\end{equation}
Here, the full covariance matrix $\Sigma$ is explicitly utilised to guarantee a unique geometric representation during comparison.

\section{Experiments}
\label{sec:experiments}

\subsection{Experimental setup}

\subsubsection{Entropy coding and datasets}

To validate the proposed non-uniform quantisation pipeline, our methods are integrated into the MPEG 3DGS-PCC test platform in accordance with the Common Test Conditions (CTC).
Our framework is evaluated using two standardised entropy codecs: G-PCC (release\_mpeg151), utilising octree-based geometry and Region Adaptive Hierarchical Transform (RAHT) attribute coding, and V-PCC (mpeg152-gs-anchor), leveraging the HEVC Test Model (HEVC Main 10 using HM-16.20) for patch-based video encoding. Note that the use of reference software leads to longer processing times compared to real-time, hardware-accelerated codecs. Testing is conducted across four standardised ratepoints to assess performance across a diverse range of bandwidth constraints. While each ratepoint utilises a distinct configuration for the entropy codecs, the preprocessing pipeline remains identical across all ratepoints. 

We evaluate our method on the official MPEG 3DGS datasets, which consist of two categories: large-scale, front-facing scenes, and high-detail captures of static objects and people (e.g., bartender\_stable, cinema\_stable, lego\_bugatti, plant\dots). Because our method incorporates a machine learning component for importance approximation, we adopt a leave-one-out strategy; one scene from each dataset is excluded for neural network training, while the remaining scenes are reserved exclusively for evaluation.

\subsubsection{Quantisation and baselines}

We compare our approach against current state-of-the-art methods, including uniform quantisation, adaptive voxelization~\cite{wang_adaptive_2025}, and FlexGaussian~\cite{tian_flexgaussian_2025}. Notably, because FlexGaussian was not originally designed for lossy entropy coding, it could not be natively integrated into the MPEG entropy codecs. Consequently, we evaluate it in its original form without modification, which yields only a single ratepoint for comparison. For the standard baselines, we allocate 18 bits to quantise position information, and 12 and 10 bits for attribute compression in G-PCC and V-PCC, respectively. The exception to this setup is adaptive voxelization~\cite{wang_adaptive_2025}, where the number of voxelisation bits is adaptively determined, and the voxelisation threshold parameter is set to 20.

\subsubsection{Evaluation metrics}

To measure the reconstruction distortion between images rendered by the 3DGS model before and after compression, we employ PSNR, SSIM, IVSSIM, and LPIPS~\cite{zhang_unreasonable_2018} as our primary synthesis quality metrics. To quantify rate-distortion (R-D) performance, we compute the Bjøntegaard Delta-Rate (BD-rate), using the uniform quantisation configurations provided by the G-PCC and V-PCC standards as anchors. Where necessary, we also provide full R-D curves for comprehensive analysis. Finally, computational complexity is evaluated by comparing execution times across the pre-processing, encoding, decoding, and post-processing stages.

\subsubsection{Implementation details}

All experiments were conducted on a workstation equipped with an Intel Core i9-13900 CPU and an NVIDIA RTX 4090 GPU. In our proposed pipeline, the number of voxelisation bits is set to 14 for front-facing scenes and 11 for object captures. For the recursive quantisation scheme, the bit-depth for the second quantisation pass is set to 2 bits for both positions and attributes. Gaussian merging is performed via weighted parameter averaging, where the weights are dynamically predicted by a tiny MLP. This MLP consists of two hidden layers of size 16 and 6, respectively, totalling 205 trainable parameters. The network was trained using the Adam optimiser with a learning rate of 0.01 for 50 epochs.

\subsection{Results}

\begin{figure}[t]
    \centering
    \includegraphics[width=1.0\linewidth]{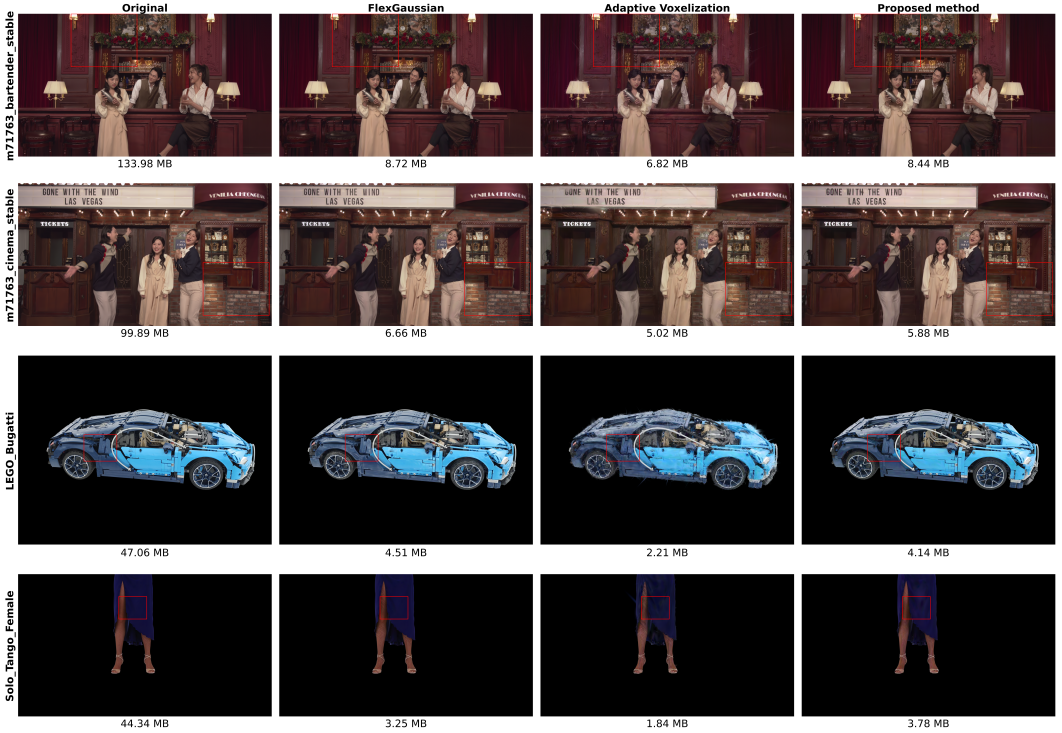}
    \caption{Qualitative comparison against the State-of-the-Art. One example scene is randomly picked per dataset, and a red box shows where our method has the lowest LPIPS compared to the other methods. The total bitrate usage is also shown below each image.}
    \Description{description}
    \label{fig:qual}
\end{figure}

\begin{figure}[t]
    \centering
    \includegraphics[width=1.0\linewidth]{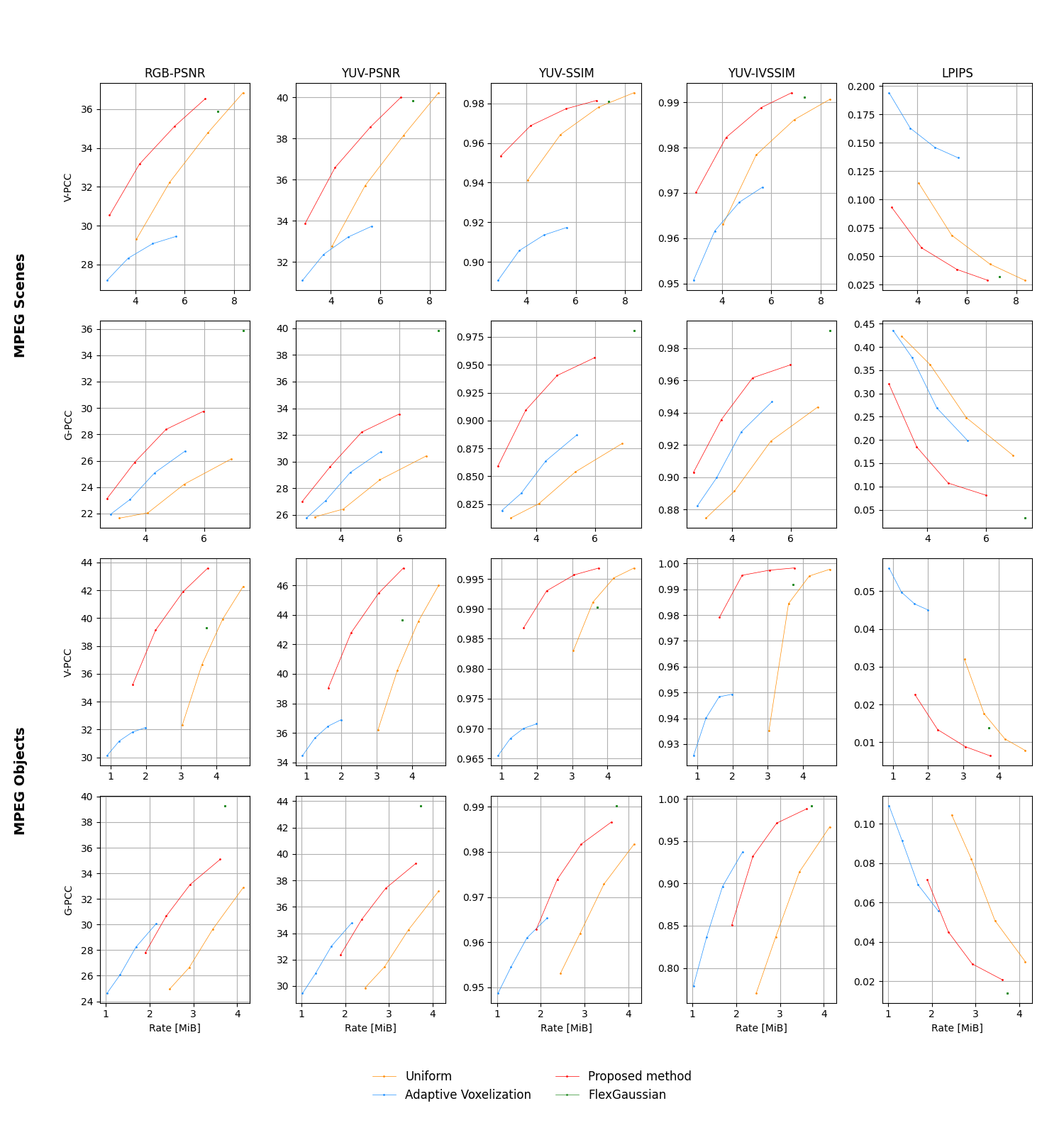}
    \caption{Rate-distortion curves comparing the proposed method against the state-of-the-art, only one rate point is given for FlexGaussian~\cite{tian_flexgaussian_2025}, because it doesn't work with lossy entropy codecs.}
    \label{fig:base_comparison}
\end{figure}

We evaluate the quantitative rate-distortion (RD) performance of the proposed method against state-of-the-art approaches in Figure~\ref{fig:base_comparison}. Our proposed method consistently outperforms uniform voxelisation across all evaluated datasets and codecs. 
Compared to Adaptive Voxelization~\cite{wang_adaptive_2025}, we achieve significantly better performance when using V-PCC and on the MPEG scenes. On the MPEG Objects dataset using G-PCC, Adaptive Voxelization exhibits better rate-distortion performance under certain configurations; however, our method yields a higher absolute quality ceiling. While FlexGaussian~\cite{tian_flexgaussian_2025} provides competitive performance at its single operational rate point, its lack of compatibility with lossy entropy codecs limits its flexibility. Currently, under specific G-PCC configurations, our method exhibits lower quality than FlexGaussian, albeit at significantly reduced bitrates. Because our framework is built on entropy-coding compatibility, we anticipate further performance gains as V-PCC and G-PCC codec configurations are progressively optimised for 3D Gaussian Splatting attributes.
A qualitative comparison is presented in Figure~\ref{fig:qual}, showing the highest-quality rate point for V-PCC. The proposed method exhibits noticeably fewer compression artefacts than FlexGaussian on the objects, while on the front-facing scenes, the subjective quality is very similar, but we save a significant amount of bitrate. Conversely, while Adaptive Voxelization yields the lowest bitrates, it introduces severe visual degradation.
Finally, we analyse the computational overhead in Table~\ref{table:base_comparison_timing}. Although our method requires a longer standalone pre-processing and quantisation time than competing approaches, this overhead becomes less pronounced when factoring in the total end-to-end entropy encoding pipeline. Crucially, in practical deployment scenarios, encoding occurs downstream of the 3DGS training phase, rendering the encoding time negligible. Meanwhile, our decoding speed remains highly competitive and closely matches uniform voxelisation, successfully shifting the computational complexity to the encoder side to ensure real-time responsiveness during playback.

\begin{table*}[t]
\caption{Comparison of the execution speed of the different approaches. The execution times of the entropy codecs are seperated from the time the quantisation and pruning takes, and is 0 for FlexGaussian which is not compatible with the lossy entropy codecs.}
\label{table:base_comparison_timing}
\centering
\resizebox{\linewidth}{!}{%
\begin{tabular}{lllrrrr}
\toprule
Dataset & Codec & Method &  Pre-Processing [s] & Encoder [s] & Decoder [s] & Post-Processing [s] \\
\midrule
\multirow[t]{8}{*}{MPEG Scenes} & \multirow[t]{4}{*}{V-PCC} & Uniform & 11.77 & 759.82 & 4.95 & 10.78 \\
 &  & FlexGaussian & 0.31 & 0.00 & 0.00 & 0.02 \\
 &  & Adaptive Voxelization & 19.40 & 405.12 & 3.50 & 7.51 \\
 &  & Proposed method & 116.18 & 351.63 & 4.08 & 9.20 \\
\cline{2-7}
 & \multirow[t]{4}{*}{G-PCC} & Uniform & 11.86 & 33.49 & 19.73 & 13.64 \\
 &  & FlexGaussian & 0.31 & 0.00 & 0.00 & 0.02 \\
 &  & Adaptive Voxelization & 18.54 & 36.34 & 22.52 & 9.43 \\
 &  & Proposed method & 112.04 & 30.11 & 18.95 & 11.27 \\
\cline{1-7} \cline{2-7}
\multirow[t]{8}{*}{MPEG Objects} & \multirow[t]{4}{*}{V-PCC} & Uniform & 4.57 & 205.74 & 4.11 & 4.21 \\
 &  & FlexGaussian & 5.74 & 0.00 & 0.00 & 0.04 \\
 &  & Adaptive Voxelization & 6.09 & 89.23 & 1.29 & 2.82 \\
 &  & Proposed method & 15.47 & 130.37 & 1.89 & 4.46 \\
\cline{2-7}
 & \multirow[t]{4}{*}{G-PCC} & Uniform & 4.53 & 21.32 & 14.36 & 5.55 \\
 &  & FlexGaussian & 22.91 & 0.00 & 0.00 & 0.05 \\
 &  & Adaptive Voxelization & 6.00 & 9.10 & 6.30 & 3.55 \\
 &  & Proposed method & 13.72 & 15.82 & 11.39 & 5.74 \\
\cline{1-7} \cline{2-7}
\bottomrule
\end{tabular}
}
\end{table*}

\section{Ablation studies}
\label{sec:abl}

In this section we provide thorough ablation studies for all the components of the proposed method.

\subsection{Voxelisation bits}

\begin{table*}[t]
\caption{Ablation study on the number of bits used for voxelisation.}
\label{table:voxelisation_bits}
\centering
\resizebox{\linewidth}{!}{%
\begin{tabular}{llllllll}
\toprule
\multirow[t]{2}{*}{Dataset} & \multirow[t]{2}{*}{Codec} & \multirow[t]{2}{*}{Bits} & \multicolumn{5}{c}{BD-Rate [\%]} \\
\cline{4-8}\\
 &  &  & RGB-PSNR & YUV-PSNR & YUV-SSIM & YUV-IVSSIM & LPIPS  \\
\midrule
\multirow[t]{8}{*}{MPEG Scenes} & \multirow[t]{4}{*}{V-PCC} & 18 & -5.63 & -4.44 & -7.93 & -0.66 & -4.98 \\
 &  & 16 & -14.44 & -13.55 & -18.51 & -10.65 & -16.26 \\
 &  & 14 & \textbf{-28.27} & -28.04 & \textbf{-27.65} & -31.54 & \textbf{-30.01} \\
 &  & 12 & 5.75 & \textbf{-30.60} & nan & \textbf{-54.96} & -17.18 \\
\cline{2-8}
 & \multirow[t]{4}{*}{G-PCC} & 18 & -11.78 & 1.91 & -29.31 & -6.33 & -8.25 \\
 &  & 16 & -23.18 & -13.82 & -43.56 & -20.79 & -24.39 \\
 &  & 14 & -48.08 & -40.78 & -59.72 & -44.50 & -44.14 \\
 &  & 12 & \textbf{-71.96} & \textbf{-69.30} & \textbf{-75.56} & \textbf{-71.51} & \textbf{-70.12} \\
\cline{1-8} \cline{2-8}
\multirow[t]{8}{*}{MPEG Objects} & \multirow[t]{4}{*}{V-PCC} & 12 & -43.38 & -43.13 & -40.72 & -45.72 & -41.89 \\
 &  & 11 & \textbf{-44.17} & -43.92 & \textbf{-41.68} & -48.51 & \textbf{-43.44} \\
 &  & 10 & -43.73 & -43.23 & -40.42 & -51.14 & -43.22 \\
 &  & 9 & -43.82 & \textbf{-44.33} & -20.26 & \textbf{-54.57} & -42.83 \\
\cline{2-8}
 & \multirow[t]{4}{*}{G-PCC} & 12 & -28.76 & -27.98 & -26.33 & -29.53 & -28.69 \\
 &  & 11 & -32.95 & -32.13 & -30.55 & -34.29 & -32.97 \\
 &  & 10 & -37.36 & -36.72 & -34.23 & -39.07 & -37.90 \\
 &  & 9 & \textbf{-43.53} & \textbf{-43.08} & \textbf{-38.94} & \textbf{-47.08} & \textbf{-45.06} \\
\cline{1-8} \cline{2-8}
\bottomrule
\end{tabular}
}
\end{table*}

\begin{figure}[t]
    \centering
    \includegraphics[width=1.0\linewidth]{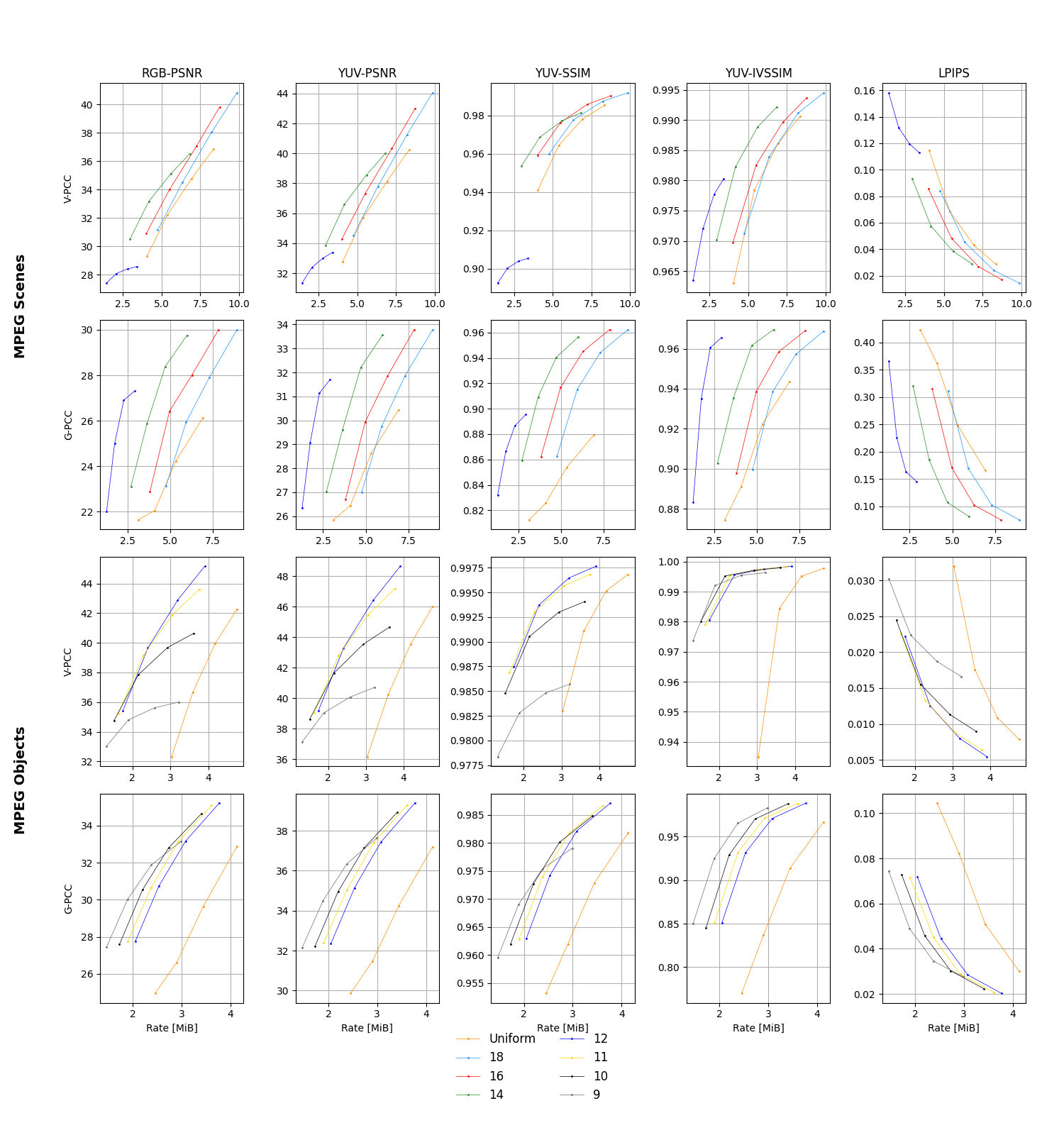}
    \caption{Rate-distortion curves showing the impact of the number of bits used for geometry quantisation for the proposed method.}
    \label{fig:vox_bits_comparison}
\end{figure}

The bit depth allocated for geometry quantisation is a critical tunable parameter in our pipeline. Per MPEG guidelines, the standard anchor for uniform quantisation is set to 18 bits, as even marginal geometry errors in 3DGS can manifest as severe visual artefacts. However, our non-uniform approach leverages the statistical distribution of the Gaussians to maintain comparable visual fidelity at a reduced bit depth. Lowering the quantisation bit depth serves a dual purpose: it directly reduces the geometry bitrate and facilitates a more aggressive merging of proximal Gaussians. As evidenced by the results in Table~\ref{table:voxelisation_bits} and Figure~\ref{fig:vox_bits_comparison}, decreasing this value results in a substantial reduction in the total bitrate. Nevertheless, there is a clear threshold below which visual quality degrades sharply; therefore, selecting a good value for the number of voxelisation bits is important to find the desired balanced between rate and distortion. We selected the number of voxelisation bits that result in a comprable reconstruction quality to uniform voxelisation.

\subsection{Weighting}

\begin{table*}[t]
\caption{Ablation study on the weighting used for non-uniform quantisation and merging.}
\label{table:weighting_comparison}
\centering
\resizebox{\linewidth}{!}{%
\begin{tabular}{llllllll}
\toprule
\multirow[t]{2}{*}{Dataset} & \multirow[t]{2}{*}{Codec} & \multirow[t]{2}{*}{Weighting} & \multicolumn{5}{c}{BD-Rate [\%]} \\
\cline{4-8}\\
 &  &  & RGB-PSNR & YUV-PSNR & YUV-SSIM & YUV-IVSSIM & LPIPS  \\
\midrule
\multirow[t]{6}{*}{MPEG Scenes} & \multirow[t]{3}{*}{V-PCC} & Uniform & -11.42 & -11.24 & -10.93 & -14.96 & -14.50 \\
 &  & Rendered & -21.07 & -20.42 & -18.49 & -24.66 & -22.15 \\
 &  & Approximate & \textbf{-28.27} & \textbf{-28.04} & \textbf{-27.65} & \textbf{-31.54} & \textbf{-30.01} \\
\cline{2-8}
 & \multirow[t]{3}{*}{G-PCC} & Uniform & -29.74 & -22.89 & -52.38 & -28.12 & -35.45 \\
 &  & Rendered & -38.14 & -30.14 & -56.08 & -34.36 & -38.83 \\
 &  & Approximate & \textbf{-48.08} & \textbf{-40.78} & \textbf{-59.72} & \textbf{-44.50} & \textbf{-44.14} \\
\cline{1-8} \cline{2-8}
\multirow[t]{6}{*}{MPEG Objects} & \multirow[t]{3}{*}{V-PCC} & Uniform & -43.78 & -43.54 & -41.32 & -47.99 & -43.24 \\
 &  & Rendered & \textbf{-44.26} & \textbf{-43.95} & -41.37 & -48.39 & -43.17 \\
 &  & Approximate & -44.17 & -43.92 & \textbf{-41.68} & \textbf{-48.51} & \textbf{-43.44} \\
\cline{2-8}
 & \multirow[t]{3}{*}{G-PCC} & Uniform & -32.43 & -31.56 & -30.11 & -33.47 & -32.52 \\
 &  & Rendered & -32.87 & -32.08 & -30.50 & -33.22 & \textbf{-33.02} \\
 &  & Approximate & \textbf{-32.95} & \textbf{-32.13} & \textbf{-30.55} & \textbf{-34.29} & -32.97 \\
\cline{1-8} \cline{2-8}
\bottomrule
\end{tabular}
}
\end{table*}

\begin{table*}[t]
\label{table:weight_comparison_timing}
\caption{Execution time comparison for the ablation study on different importance weighting approaches.}
\centering
\resizebox{\linewidth}{!}{%
\begin{tabular}{lllrrrr}
\toprule
Dataset & Codec & Method &  Pre-Processing [s] & Encoder [s] & Decoder [s] & Post-Processing [s] \\
\midrule
\multirow[t]{6}{*}{MPEG Scenes} & \multirow[t]{3}{*}{V-PCC} & Uniform & 72.67 & 479.49 & 4.98 & 11.37 \\
 &  & Rendered & 324.26 & 419.56 & 4.51 & 9.97 \\
 &  & Approximate & 116.18 & 351.63 & 4.08 & 9.20 \\
\cline{2-7}
 & \multirow[t]{3}{*}{G-PCC} & Uniform & 67.02 & 34.84 & 21.25 & 14.98 \\
 &  & Rendered & 330.08 & 31.57 & 19.98 & 13.48 \\
 &  & Approximate & 112.04 & 30.11 & 18.95 & 11.27 \\
\cline{1-7} \cline{2-7}
\multirow[t]{6}{*}{MPEG Objects} & \multirow[t]{3}{*}{V-PCC} & Uniform & 15.11 & 134.51 & 1.98 & 4.45 \\
 &  & Rendered & 375.72 & 131.65 & 1.95 & 4.36 \\
 &  & Approximate & 15.47 & 130.37 & 1.89 & 4.46 \\
\cline{2-7}
 & \multirow[t]{3}{*}{G-PCC} & Uniform & 23.13 & 26.57 & 16.05 & 8.82 \\
 &  & Rendered & 2131.01 & 26.80 & 16.34 & 8.66 \\
 &  & Approximate & 13.72 & 15.82 & 11.39 & 5.74 \\
\cline{1-7} \cline{2-7}
\bottomrule
\end{tabular}
}
\end{table*}

The proposed quantisation and merging frameworks both use importance weighting to prioritise important Gaussians. We evaluated three distinct weighting configurations: a rendering-based approach that minimises MSE by calculating the direct contribution of each Gaussian to the training views, a learned MLP-based proxy designed to approximate these rendering-based weights, and a uniform weighting baseline to isolate the impact of the importance metrics. As shown in Table~\ref{table:weighting_comparison}, importance weighting provides consistent and significant gains in compression efficiency on the MPEG scenes, while the gains are smaller for the objects. From a computational perspective, the rendering-based approach introduces a substantial increase in pre-processing time. Conversely, the MLP-based approximation achieves similar, if not better rate-distortion performance while incurring negligible computational overhead. This makes the MLP approach the most viable candidate for real-time standardisation workflows. In addition, the MLP is very small and generalises easily to new scenes, as it is designed to be a simple rule.

\subsection{Merging}

\begin{table*}[t]
\caption{Ablation study on the different proposed methods to merge co-located primitives.}
\label{table:merge_comparison}
\centering
\resizebox{\linewidth}{!}{%
\begin{tabular}{llllllll}
\toprule
\multirow[t]{2}{*}{Dataset} & \multirow[t]{2}{*}{Codec} & \multirow[t]{2}{*}{Merging} & \multicolumn{5}{c}{BD-Rate [\%]} \\
\cline{4-8}\\
 &  &  & RGB-PSNR & YUV-PSNR & YUV-SSIM & YUV-IVSSIM & LPIPS  \\
\midrule
\multirow[t]{10}{*}{MPEG Scenes} & \multirow[t]{5}{*}{V-PCC} & No Merging & -10.70 & -10.25 & -11.30 & -13.34 & -14.82 \\
 &  & Basic Averaging & -9.08 & -11.03 & -17.37 & -20.93 & -16.25 \\
 &  & Weighted Parameter Averaging & \textbf{-28.27} & \textbf{-28.04} & \textbf{-27.65} & \textbf{-31.54} & \textbf{-30.01} \\
 &  & Weighted Covariance & -6.26 & -8.26 & -14.96 & -21.50 & -2.94 \\
 &  & Weighted Log-Euclidean & -13.62 & -14.73 & -18.54 & -25.40 & -10.44 \\
\cline{2-8}
 & \multirow[t]{5}{*}{G-PCC} & No Merging & -24.05 & -19.86 & -52.36 & -22.79 & -34.84 \\
 &  & Basic Averaging & -40.24 & -34.50 & -59.50 & -38.04 & -42.37 \\
 &  & Weighted Parameter Averaging & \textbf{-48.08} & \textbf{-40.78} & \textbf{-59.72} & \textbf{-44.50} & \textbf{-44.14} \\
 &  & Weighted Covariance & -43.66 & -36.53 & -56.84 & -40.83 & -37.98 \\
 &  & Weighted Log-Euclidean & -41.86 & -35.06 & -56.32 & -40.37 & -37.96 \\
\cline{1-8} \cline{2-8}
\multirow[t]{10}{*}{MPEG Objects} & \multirow[t]{5}{*}{V-PCC} & No Merging & -43.88 & -43.66 & -41.19 & -48.07 & -43.21 \\
 &  & Basic Averaging & -41.59 & -41.36 & -40.97 & -47.67 & -42.23 \\
 &  & Weighted Parameter Averaging & \textbf{-44.17} & \textbf{-43.92} & \textbf{-41.68} & \textbf{-48.51} & \textbf{-43.44} \\
 &  & Weighted Covariance & -41.00 & -40.53 & -39.68 & -45.57 & -39.89 \\
 &  & Weighted Log-Euclidean & -41.16 & -40.74 & -39.79 & -45.42 & -40.02 \\
\cline{2-8}
 & \multirow[t]{5}{*}{G-PCC} & No Merging & -32.60 & -31.70 & -30.09 & -33.85 & -32.62 \\
 &  & Basic Averaging & -32.47 & -31.65 & -30.35 & -33.53 & -32.73 \\
 &  & Weighted Parameter Averaging & \textbf{-32.95} & \textbf{-32.13} & \textbf{-30.55} & \textbf{-34.29} & \textbf{-32.97} \\
 &  & Weighted Covariance & -31.03 & -30.34 & -29.00 & -32.36 & -31.18 \\
 &  & Weighted Log-Euclidean & -31.23 & -30.51 & -29.11 & -32.39 & -31.32 \\
\cline{1-8} \cline{2-8}
\bottomrule
\end{tabular}
}
\end{table*}

\begin{table*}[t]
\caption{Ablation study on the dissimilarity metric used as a criterion for merging co-located Gaussians.}
\label{table:merge_criterion}
\centering
\resizebox{\linewidth}{!}{%
\begin{tabular}{lllllllll}
\toprule
 \multirow[t]{2}{*}{Dataset} & \multirow[t]{2}{*}{Codec} & \multirow[t]{2}{*}{Merging} & \multirow[t]{2}{*}{\# Splats}& \multicolumn{5}{c}{BD-Rate [\%]} \\
\cline{5-9}\\
 & &  &  & RGB-PSNR & YUV-PSNR & YUV-SSIM & YUV-IVSSIM & LPIPS  \\
\midrule
\multirow[t]{6}{*}{MPEG Scenes} & \multirow[t]{3}{*}{V-PCC} & No merging & 495486.5 & -10.70 & -10.25 & -11.30 & -13.34 & -14.82 \\
 &  & With Criterion & 376527.5 & \textbf{-28.27} & \textbf{-28.04} & \textbf{-27.65} & -31.54 & \textbf{-30.01} \\
 &  & Merging All & 375847.5 & -27.86 & -27.86 & -27.55 & \textbf{-31.78} & -29.72 \\
\cline{2-9}
 & \multirow[t]{3}{*}{G-PCC} & No merging & 495486.5 & -24.06 & -19.87 & -52.35 & -22.80 & -34.83 \\
 &  & With Criterion & 376527.5 & \textbf{-48.08} & \textbf{-40.78} & -59.72 & \textbf{-44.50} & \textbf{-44.14} \\
 &  & Merging All & 375847.5 & -44.21 & -36.91 & \textbf{-60.06} & -40.65 & -43.68 \\
\cline{1-9} \cline{2-9}
\multirow[t]{6}{*}{MPEG Objects} & \multirow[t]{3}{*}{V-PCC} & No merging & 193330.25 & -43.88 & -43.66 & -41.19 & -48.07 & -43.21 \\
 &  & With Criterion & 189955.0 & \textbf{-44.17} & \textbf{-43.92} & \textbf{-41.68} & -48.51 & \textbf{-43.44} \\
 &  & Merging All & 189935.0 & -44.08 & -43.83 & -41.66 & \textbf{-48.83} & -43.33 \\
\cline{2-9}
 & \multirow[t]{3}{*}{G-PCC} & No merging & 193330.25 & -32.60 & -31.69 & -30.09 & -33.85 & -32.62 \\
 &  & With Criterion & 189955.0 & -32.95 & \textbf{-32.13} & \textbf{-30.55} & \textbf{-34.29} & -32.97 \\
 &  & Merging All & 189935.0 & \textbf{-32.95} & -32.10 & -30.52 & -34.24 & \textbf{-33.00} \\
\cline{1-9} \cline{2-9}
\bottomrule
\end{tabular}
}
\end{table*}

In this subsection, we compare the performance of the three different merging approaches, as discussed in Section~\ref{subsec:merging}, and compare them to not merging, or a more basic averaging, similar to Adaptive Voxelization~\cite{wang_adaptive_2025}. The direct parameter averaging clearly outperforms all other approaches, proving to be more accurate than the covariance-based averaging. Additionally, the results demonstrate the impact of the importance weighting, in most cases the basic averaging is worse than not merging at all.

\subsection{Second quantisation bits}

\begin{table*}[t]
\caption{Ablation study on the effect of the number of quantisation bits for the second quantisation application.}
\label{table:qbits}
\centering
\resizebox{\linewidth}{!}{%
\begin{tabular}{llllllll}
\toprule
\multirow[t]{2}{*}{Dataset} & \multirow[t]{2}{*}{Codec} & \multirow[t]{2}{*}{bits} & \multicolumn{5}{c}{BD-Rate [\%]} \\
\cline{4-8}\\
 &  &  & RGB-PSNR & YUV-PSNR & YUV-SSIM & YUV-IVSSIM & LPIPS  \\
\midrule
\multirow[t]{8}{*}{MPEG Scenes} & \multirow[t]{4}{*}{V-PCC} & 1 & -24.78 & -25.09 & \textbf{-27.67} & -29.72 & -28.73 \\
 &  & 2 & \textbf{-28.27} & \textbf{-28.04} & -27.65 & \textbf{-31.54} & \textbf{-30.01} \\
 &  & 3 & -27.27 & -26.70 & -27.31 & -30.15 & -29.60 \\
 &  & 4 & -27.89 & -27.68 & -27.23 & -31.14 & -29.86 \\
\cline{2-8}
 & \multirow[t]{4}{*}{G-PCC} & 1 & -45.68 & -37.45 & \textbf{-60.35} & -40.42 & -44.12 \\
 &  & 2 & \textbf{-48.08} & \textbf{-40.78} & -59.72 & \textbf{-44.50} & \textbf{-44.14} \\
 &  & 3 & -43.82 & -35.00 & -59.74 & -40.56 & -43.57 \\
 &  & 4 & -45.11 & -37.25 & -59.42 & -43.84 & -43.62 \\
\cline{1-8} \cline{2-8}
\multirow[t]{8}{*}{MPEG Objects} & \multirow[t]{4}{*}{V-PCC} & 1 & -43.45 & -43.22 & \textbf{-42.71} & -48.05 & \textbf{-44.19} \\
 &  & 2 & -44.17 & -43.92 & -41.68 & \textbf{-48.51} & -43.44 \\
 &  & 3 & \textbf{-44.25} & \textbf{-43.95} & -41.48 & -48.31 & -43.52 \\
 &  & 4 & -44.21 & -43.92 & -41.57 & -48.19 & -43.59 \\
\cline{2-8}
 & \multirow[t]{4}{*}{G-PCC} & 1 & \textbf{-33.00} & \textbf{-32.24} & \textbf{-30.63} & -34.20 & \textbf{-33.05} \\
 &  & 2 & -32.95 & -32.13 & -30.55 & \textbf{-34.29} & -32.97 \\
 &  & 3 & -32.84 & -32.10 & -30.45 & -33.96 & -32.90 \\
 &  & 4 & -32.60 & -31.78 & -30.33 & -33.51 & -32.72 \\
\cline{1-8} \cline{2-8}
\bottomrule
\end{tabular}
}
\end{table*}

The number of quantisation bits for the seconds stage quantisation, similarly to the first, is a parameter that allows to balance reconstruction quality and bitrate. We experimented with different values and observe that it is possible to get very high quality reconstructions with a small value like 2 (Table~\ref{table:qbits}). 

\subsection{Attributes}

The proposed method can be applied to the positions and all attributes of a learned 3dgs scene. In this last ablation study we investigate enabling the non-uniform quantisation only on one attribute at a time. The results in Table~\ref{table:abl_attr} show that non-uniform quantisation of the positions consistently brings the largest gains. The gains and losses for other attributes using V-PCC are much smaller, while they are more random for G-PCC.

\begin{table*}[t]
\caption{Ablation study on performance per attribute, gains are marked in \textbf{bold}}
\label{table:abl_attr}
\centering
\resizebox{\linewidth}{!}{%
\begin{tabular}{llllllll}
\toprule
\multirow[t]{2}{*}{Dataset} & \multirow[t]{2}{*}{Codec} & \multirow[t]{2}{*}{Attribute} & \multicolumn{5}{c}{BD-Rate [\%]} \\
\cline{4-8}\\
 &  &  & RGB-PSNR & YUV-PSNR & YUV-SSIM & YUV-IVSSIM & LPIPS  \\
\midrule
\multirow[t]{10}{*}{MPEG Scenes} & \multirow[t]{5}{*}{V-PCC} & Positions & \textbf{-22.30} & \textbf{-22.71} & \textbf{-23.24} & \textbf{-29.73} & \textbf{-25.16} \\
 &  & Spherical Harmonics & \textbf{-6.60} & \textbf{-6.53} & \textbf{-0.95} & \textbf{-3.41} & \textbf{-4.46} \\
 &  & Scales & 0.78 & 0.42 & \textbf{-0.22} & 0.33 & \textbf{-0.04} \\
 &  & Rotations & 1.36 & 1.22 & \textbf{-7.47} & 2.67 & \textbf{-1.87} \\
 &  & Opacities & \textbf{-0.11} & \textbf{-0.50} & \textbf{-1.35} & \textbf{-0.35} & \textbf{-1.23} \\
\cline{2-8}
 & \multirow[t]{5}{*}{G-PCC} & Positions & \textbf{-37.68} & \textbf{-33.87} & \textbf{-55.73} & \textbf{-35.50} & \textbf{-28.25} \\
 &  & Spherical Harmonics & 6.14 & 11.78 & \textbf{-28.94} & 9.44 & 14.62 \\
 &  & Scales & 9.30 & 16.80 & \textbf{-27.85} & 12.77 & 16.62 \\
 &  & Rotations & \textbf{-32.74} & \textbf{-15.34} & \textbf{-25.77} & \textbf{-21.82} & \textbf{-16.23} \\
 &  & Opacities & 14.81 & 17.91 & \textbf{-26.69} & 18.71 & 17.83 \\
\cline{1-8} \cline{2-8}
\multirow[t]{10}{*}{MPEG Objects} & \multirow[t]{5}{*}{V-PCC} & Positions & \textbf{-44.79} & \textbf{-44.60} & \textbf{-44.37} & \textbf{-50.73} & \textbf{-46.05} \\
 &  & Spherical Harmonics & \textbf{-1.05} & \textbf{-0.74} & 0.89 & \textbf{-0.22} & 0.96 \\
 &  & Scales & \textbf{-0.13} & \textbf{-0.12} & \textbf{-0.06} & 0.19 & 0.14 \\
 &  & Rotations & 0.94 & 0.68 & \textbf{-0.39} & 2.34 & 0.85 \\
 &  & Opacities & \textbf{-0.73} & \textbf{-0.70} & \textbf{-0.58} & \textbf{-0.44} & \textbf{-0.44} \\
\cline{2-8}
 & \multirow[t]{5}{*}{G-PCC} & Positions & \textbf{-21.79} & \textbf{-21.81} & \textbf{-22.20} & \textbf{-22.73} & \textbf{-22.52} \\
 &  & Spherical Harmonics & 9.85 & 9.90 & 10.48 & 11.03 & 10.76 \\
 &  & Scales & 10.06 & 10.10 & 10.50 & 11.47 & 10.79 \\
 &  & Rotations & \textbf{-3.97} & \textbf{-2.74} & \textbf{-3.44} & \textbf{-3.21} & \textbf{-4.58} \\
 &  & Opacities & 8.69 & 8.73 & 10.95 & 10.59 & 9.98 \\
\cline{1-8} \cline{2-8}
\bottomrule
\end{tabular}
}
\end{table*}

\section{Discussion}
\label{sec:discussion}

Across the evaluated datasets, the proposed approach consistently outperforms the currently adopted uniform quantisation baseline. Our method achieves significant bitrate savings by strategically reducing the quantisation bit depth, while simultaneously improving reconstruction quality by allocating the available bits where they have the highest perceptual impact. 
Regarding computational overhead, the proposed pipeline introduces an asymmetric complexity trade-off. While the additional logic increases the standalone pre-processing time, the overall encoding pipeline remains highly manageable. In fact, for specific configurations, the total encoding duration decreases due to the substantial reduction in primitive count achieved by our weighted merging stage. Crucially, the client-side decoding and post-processing times remain entirely comparable to uniform quantisation, ensuring that real-time rendering and system responsiveness are not compromised. Consequently, our framework offers a compelling, high-performance alternative to uniform quantisation for the upcoming MPEG 3DGS compression standards.
Furthermore, the proposed weighted merging strategy acts as an effective, lightweight alternative to standard pruning techniques. Rather than discarding low-impact primitives, it targets structural redundancy by consolidating co-located Gaussians only when they fall below a strict dissimilarity threshold. This underlying principle is fully complementary to existing pruning algorithms and can be integrated into broader compression workflows. Finally, the capacity to modulate the quantisation and merging pipelines via importance weighting extends beyond minimising average compression distortion; it provides a flexible foundation for application-specific paradigms, such as prioritizing the reconstruction quality of semantically critical objects within a scene.

\section{Conclusion}
\label{sec:conclusion}

In this paper, we presented a comprehensive, non-uniform quantisation pipeline engineered specifically for the compression of 3D Gaussian Splatting (3DGS) models. By integrating an importance-weighting framework with a recursive weighted Lloyd-Max quantiser and advanced structural merging techniques, our approach successfully achieves state-of-the-art rate-distortion performance. 
The proposed methods are entirely agnostic to the underlying entropy coding engine and remain compatible with any point-cloud-based splatting representation. Developed as formal contributions to the upcoming G-PCC and V-PCC based MPEG 3DGS standardization tracks, our framework has demonstrated immediate industrial relevance, with the proposed weighted merging method already being officially adopted into the V-PCC Amd1 for GS.

\bibliographystyle{ACM-Reference-Format}
\bibliography{references}

@article{isoiec_isoiec_2025,
	title = {{ISO}/{IEC} 23090-5:2025 {Part}: {Visual} volumetric video-based coding ({V3C}) and video-based point cloud compression ({V}-{PCC})},
	author = {{ISO/IEC}},
	year = {2025},
}

@article{isoiec_isoiec_2023,
	title = {{ISO}/{IEC} 23090-9 {Geometry}-based {Point} {Cloud} {Coding} ({GPCC}):2023},
	author = {{ISO/IEC}},
	year = {2023},
}

@article{zhu_low-rank_2026,
	title = {Low-{Rank} {Approximation} for {Efficient} {Compression} of {Gaussian} {Splatting} {Spherical} {Harmonics}},
	issn = {1558-2205},
	url = {https://ieeexplore.ieee.org/abstract/document/11419127},
	doi = {10.1109/TCSVT.2026.3670178},
	urldate = {2026-06-11},
	journal = {IEEE Transactions on Circuits and Systems for Video Technology},
	author = {Zhu, Zhiwei and Li, Sicheng and Liao, Yiyi and Yu, Lu},
	year = {2026},
	pages = {1--1},
}

@inproceedings{fang_mini-splatting_2024,
	address = {Cham},
	title = {Mini-{Splatting}: {Representing} {Scenes} with a {Constrained} {Number} of {Gaussians}},
	isbn = {978-3-031-72980-5},
	shorttitle = {Mini-{Splatting}},
	doi = {10.1007/978-3-031-72980-5_10},
	language = {en},
	booktitle = {Computer {Vision} – {ECCV} 2024},
	publisher = {Springer Nature Switzerland},
	author = {Fang, Guangchi and Wang, Bing},
	editor = {Leonardis, Aleš and Ricci, Elisa and Roth, Stefan and Russakovsky, Olga and Sattler, Torsten and Varol, Gül},
	year = {2024},
	pages = {165--181},
}

@inproceedings{lee_compact_2024,
	title = {Compact {3D} {Gaussian} {Representation} for {Radiance} {Field}},
	url = {https://openaccess.thecvf.com/content/CVPR2024/html/Lee_Compact_3D_Gaussian_Representation_for_Radiance_Field_CVPR_2024_paper.html},
	language = {en},
	urldate = {2026-06-11},
	author = {Lee, Joo Chan and Rho, Daniel and Sun, Xiangyu and Ko, Jong Hwan and Park, Eunbyung},
	year = {2024},
	pages = {21719--21728},
}

@inproceedings{zhang_unreasonable_2018,
	title = {The {Unreasonable} {Effectiveness} of {Deep} {Features} as a {Perceptual} {Metric}},
	url = {https://openaccess.thecvf.com/content_cvpr_2018/html/Zhang_The_Unreasonable_Effectiveness_CVPR_2018_paper.html},
	urldate = {2026-06-09},
	author = {Zhang, Richard and Isola, Phillip and Efros, Alexei A. and Shechtman, Eli and Wang, Oliver},
	year = {2018},
	pages = {586--595},
}

@book{gersho_vector_1992,
	address = {Boston, MA},
	title = {Vector {Quantization} and {Signal} {Compression}},
	copyright = {http://www.springer.com/tdm},
	isbn = {978-1-4613-6612-6 978-1-4615-3626-0},
	url = {http://link.springer.com/10.1007/978-1-4615-3626-0},
	urldate = {2026-06-04},
	publisher = {Springer US},
	author = {Gersho, Allen and Gray, Robert M.},
	year = {1992},
	doi = {10.1007/978-1-4615-3626-0},
}

@inproceedings{bang_mpeg_2026,
	title = {{MPEG} {Explorations} {Toward} {3D} {Gaussian} {Splat} {Coding} and {Standardization}},
	url = {https://ieeexplore.ieee.org/abstract/document/11510414},
	doi = {10.1109/DCC66757.2026.00045},
	urldate = {2026-05-21},
	booktitle = {2026 {Data} {Compression} {Conference} ({DCC})},
	author = {Bang, Gun and Liao, Yiyi and Zaghetto, Alexandre and Preda, Marius and Yu, Lu},
	month = mar,
	year = {2026},
	note = {ISSN: 2375-0359},
	pages = {372--381},
}

@inproceedings{alface_coding_2025,
	title = {Coding {Gaussian} {Splat} {Scenes} with {V3C}/{V}-{PCC}},
	url = {https://ieeexplore.ieee.org/abstract/document/11366782},
	doi = {10.1109/ISM66958.2025.00021},
	urldate = {2026-05-21},
	booktitle = {2025 {International} {Symposium} on {Multimedia} ({ISM})},
	author = {Alface, Patrice Rondao and Ilola, Lauri and Kondrad, Lukasz},
	month = dec,
	year = {2025},
	pages = {64--68},
}

@misc{jurca_fourier_2026,
	title = {Fourier {Splatting}: {Generalized} {Fourier} encoded primitives for scalable radiance fields},
	shorttitle = {Fourier {Splatting}},
	url = {http://arxiv.org/abs/2603.19834},
	doi = {10.48550/arXiv.2603.19834},
	urldate = {2026-05-21},
	publisher = {arXiv},
	author = {Jurca, Mihnea-Bogdan and hauwermeiren, Bert Van and Munteanu, Adrian},
	month = apr,
	year = {2026},
	note = {arXiv:2603.19834 [cs.CV]},
}

@misc{held_triangle_2025,
	title = {Triangle {Splatting} for {Real}-{Time} {Radiance} {Field} {Rendering}},
	url = {http://arxiv.org/abs/2505.19175},
	doi = {10.48550/arXiv.2505.19175},
	urldate = {2026-03-12},
	publisher = {arXiv},
	author = {Held, Jan and Vandeghen, Renaud and Deliege, Adrien and Hamdi, Abdullah and Giancola, Silvio and Cioppa, Anthony and Vedaldi, Andrea and Ghanem, Bernard and Tagliasacchi, Andrea and Droogenbroeck, Marc Van},
	month = may,
	year = {2025},
	note = {arXiv:2505.19175 [cs]},
}

@inproceedings{zhang_quadratic_2025,
	title = {Quadratic {Gaussian} {Splatting}: {High} {Quality} {Surface} {Reconstruction} with {Second}-order {Geometric} {Primitives}},
	shorttitle = {Quadratic {Gaussian} {Splatting}},
	url = {https://openaccess.thecvf.com/content/ICCV2025/html/Zhang_Quadratic_Gaussian_Splatting_High_Quality_Surface_Reconstruction_with_Second-order_Geometric_ICCV_2025_paper.html},
	language = {en},
	urldate = {2026-03-03},
	author = {Zhang, Ziyu and Huang, Binbin and Jiang, Hanqing and Zhou, Liyang and Xiang, Xiaojun and Shen, Shuhan},
	year = {2025},
	pages = {28260--28270},
}

@article{arsigny_geometric_2007,
	title = {Geometric {Means} in a {Novel} {Vector} {Space} {Structure} on {Symmetric} {Positive}‐{Definite} {Matrices}},
	volume = {29},
	url = {https://doi.org/10.1137/050637996},
	doi = {10.1137/050637996},
	language = {en},
	number = {1},
	urldate = {2026-02-09},
	journal = {SIAM Journal on Matrix Analysis and Applications},
	author = {Arsigny, Vincent and Fillard, Pierre and Pennec, Xavier and Ayache, Nicholas},
	year = {2007},
	pages = {328--347},
}

@article{markley_averaging_2007,
	title = {Averaging {Quaternions}},
	volume = {30},
	issn = {0731-5090},
	url = {https://doi.org/10.2514/1.28949},
	doi = {10.2514/1.28949},
	number = {4},
	urldate = {2026-02-05},
	journal = {Journal of Guidance, Control, and Dynamics},
	publisher = {American Institute of Aeronautics and Astronautics},
	author = {Markley, F. Landis and Cheng, Yang and Crassidis, John L. and Oshman, Yaakov},
	year = {2007},
	note = {\_eprint: https://doi.org/10.2514/1.28949},
	pages = {1193--1197},
}

@inproceedings{huang_2d_2024,
	address = {New York, NY, USA},
	series = {{SIGGRAPH} '24},
	title = {{2D} {Gaussian} {Splatting} for {Geometrically} {Accurate} {Radiance} {Fields}},
	isbn = {979-8-4007-0525-0},
	url = {https://dl.acm.org/doi/10.1145/3641519.3657428},
	doi = {10.1145/3641519.3657428},
	urldate = {2026-02-05},
	booktitle = {{ACM} {SIGGRAPH} 2024 {Conference} {Papers}},
	publisher = {Association for Computing Machinery},
	author = {Huang, Binbin and Yu, Zehao and Chen, Anpei and Geiger, Andreas and Gao, Shenghua},
	month = jul,
	year = {2024},
	pages = {1--11},
}

@inproceedings{liu_deformable_2025,
	address = {New York, NY, USA},
	series = {{SIGGRAPH} {Conference} {Papers} '25},
	title = {Deformable {Beta} {Splatting}},
	isbn = {979-8-4007-1540-2},
	url = {https://dl.acm.org/doi/10.1145/3721238.3730716},
	doi = {10.1145/3721238.3730716},
	urldate = {2026-02-05},
	booktitle = {Proceedings of the {Special} {Interest} {Group} on {Computer} {Graphics} and {Interactive} {Techniques} {Conference} {Conference} {Papers}},
	publisher = {Association for Computing Machinery},
	author = {Liu, Rong and Sun, Dylan and Chen, Meida and Wang, Yue and Feng, Andrew},
	month = jul,
	year = {2025},
	pages = {1--11},
}

@inproceedings{hamdi_ges_2024,
	title = {{GES} : {Generalized} {Exponential} {Splatting} for {Efficient} {Radiance} {Field} {Rendering}},
	shorttitle = {{GES}},
	url = {https://openaccess.thecvf.com/content/CVPR2024/html/Hamdi_GES__Generalized_Exponential_Splatting_for_Efficient_Radiance_Field_Rendering_CVPR_2024_paper.html},
	language = {en},
	urldate = {2026-02-05},
	author = {Hamdi, Abdullah and Melas-Kyriazi, Luke and Mai, Jinjie and Qian, Guocheng and Liu, Ruoshi and Vondrick, Carl and Ghanem, Bernard and Vedaldi, Andrea},
	year = {2024},
	pages = {19812--19822},
}

@article{wu_lfgs_2025,
	title = {{LFGS}: {A} lightweight framework for efficient {3D} {Gaussian} {Splatting} with minimal memory footprint},
	volume = {131},
	issn = {0097-8493},
	shorttitle = {{LFGS}},
	url = {https://www.sciencedirect.com/science/article/pii/S0097849325001505},
	doi = {10.1016/j.cag.2025.104309},
	urldate = {2026-02-05},
	journal = {Computers \& Graphics},
	author = {Wu, Ruiqi and Jiao, Linlin and Liu, Gang and Zhu, Li and Fei, Xuan and Mu, Yashuang and Fan, Chao},
	month = oct,
	year = {2025},
	pages = {104309},
}

@article{mildenhall_nerf_2021,
	title = {{NeRF}: representing scenes as neural radiance fields for view synthesis},
	volume = {65},
	issn = {0001-0782},
	shorttitle = {{NeRF}},
	url = {https://dl.acm.org/doi/10.1145/3503250},
	doi = {10.1145/3503250},
	number = {1},
	urldate = {2026-02-05},
	journal = {Commun. ACM},
	author = {Mildenhall, Ben and Srinivasan, Pratul P. and Tancik, Matthew and Barron, Jonathan T. and Ramamoorthi, Ravi and Ng, Ren},
	month = dec,
	year = {2021},
	pages = {99--106},
}

@article{van_hauwermeiren_non-uniform_2025,
	title = {Non-{Uniform} {Voxelisation} for {Point} {Cloud} {Compression}},
	volume = {25},
	copyright = {http://creativecommons.org/licenses/by/3.0/},
	issn = {1424-8220},
	url = {https://www.mdpi.com/1424-8220/25/3/865},
	doi = {10.3390/s25030865},
	language = {en},
	number = {3},
	urldate = {2026-02-05},
	journal = {Sensors},
	publisher = {Multidisciplinary Digital Publishing Institute},
	author = {Van hauwermeiren, Bert and Denis, Leon and Munteanu, Adrian},
	month = jan,
	year = {2025},
	pages = {865},
}

@article{kerbl_3d_2023,
	title = {{3D} {Gaussian} {Splatting} for {Real}-{Time} {Radiance} {Field} {Rendering}},
	volume = {42},
	issn = {0730-0301},
	url = {https://dl.acm.org/doi/10.1145/3592433},
	doi = {10.1145/3592433},
	number = {4},
	urldate = {2026-02-05},
	journal = {ACM Trans. Graph.},
	author = {Kerbl, Bernhard and Kopanas, Georgios and Leimkuehler, Thomas and Drettakis, George},
	month = jul,
	year = {2023},
	pages = {139:1--139:14},
}

@inproceedings{lu_scaffold-gs_2024,
	title = {Scaffold-{GS}: {Structured} {3D} {Gaussians} for {View}-{Adaptive} {Rendering}},
	shorttitle = {Scaffold-{GS}},
	url = {https://openaccess.thecvf.com/content/CVPR2024/html/Lu_Scaffold-GS_Structured_3D_Gaussians_for_View-Adaptive_Rendering_CVPR_2024_paper.html},
	language = {en},
	urldate = {2025-12-28},
	author = {Lu, Tao and Yu, Mulin and Xu, Linning and Xiangli, Yuanbo and Wang, Limin and Lin, Dahua and Dai, Bo},
	year = {2024},
	pages = {20654--20664},
}

@inproceedings{navaneet_compgs_2025,
	address = {Cham},
	title = {{CompGS}: {Smaller} and {Faster} {Gaussian} {Splatting} with {Vector} {Quantization}},
	isbn = {978-3-031-73411-3},
	shorttitle = {{CompGS}},
	doi = {10.1007/978-3-031-73411-3_19},
	language = {en},
	booktitle = {Computer {Vision} – {ECCV} 2024},
	publisher = {Springer Nature Switzerland},
	author = {Navaneet, K. L. and Pourahmadi Meibodi, Kossar and Abbasi Koohpayegani, Soroush and Pirsiavash, Hamed},
	editor = {Leonardis, Aleš and Ricci, Elisa and Roth, Stefan and Russakovsky, Olga and Sattler, Torsten and Varol, Gül},
	year = {2025},
	pages = {330--349},
}

@inproceedings{chen_hac_2025,
	address = {Cham},
	title = {{HAC}: {Hash}-{Grid} {Assisted} {Context} for {3D} {Gaussian} {Splatting} {Compression}},
	isbn = {978-3-031-72667-5},
	shorttitle = {{HAC}},
	doi = {10.1007/978-3-031-72667-5_24},
	language = {en},
	booktitle = {Computer {Vision} – {ECCV} 2024},
	publisher = {Springer Nature Switzerland},
	author = {Chen, Yihang and Wu, Qianyi and Lin, Weiyao and Harandi, Mehrtash and Cai, Jianfei},
	editor = {Leonardis, Aleš and Ricci, Elisa and Roth, Stefan and Russakovsky, Olga and Sattler, Torsten and Varol, Gül},
	year = {2025},
	pages = {422--438},
}

@inproceedings{wang_adaptive_2025,
	title = {Adaptive {Voxelization} for {Transform} {Coding} of {3D} {Gaussian} {Splatting} {Data}},
	issn = {2381-8549},
	url = {https://ieeexplore.ieee.org/document/11084522},
	doi = {10.1109/ICIP55913.2025.11084522},
	urldate = {2025-11-08},
	booktitle = {2025 {IEEE} {International} {Conference} on {Image} {Processing} ({ICIP})},
	author = {Wang, Chenjunjie and Sridhara, Shashank N. and Pavez, Eduardo and Ortega, Antonio and Chang, Cheng},
	month = sep,
	year = {2025},
	pages = {2414--2419},
}

@inproceedings{tian_flexgaussian_2025,
	address = {New York, NY, USA},
	series = {{MM} '25},
	title = {{FlexGaussian}: {Flexible} and {Cost}-{Effective} {Training}-{Free} {Compression} for {3D} {Gaussian} {Splatting}},
	isbn = {979-8-4007-2035-2},
	shorttitle = {{FlexGaussian}},
	url = {https://dl.acm.org/doi/10.1145/3746027.3754744},
	doi = {10.1145/3746027.3754744},
	urldate = {2025-11-08},
	booktitle = {Proceedings of the 33rd {ACM} {International} {Conference} on {Multimedia}},
	publisher = {Association for Computing Machinery},
	author = {Tian, Boyuan and Gao, Qizhe and Xianyu, Siran and Cui, Xiaotong and Zhang, Minjia},
	month = oct,
	year = {2025},
	pages = {7287--7296},
}

@inproceedings{xie_mesongs_2025,
	address = {Cham},
	title = {{MesonGS}: {Post}-training {Compression} of {3D} {Gaussians} via {Efficient} {Attribute} {Transformation}},
	isbn = {978-3-031-73414-4},
	shorttitle = {{MesonGS}},
	doi = {10.1007/978-3-031-73414-4_25},
	language = {en},
	booktitle = {Computer {Vision} – {ECCV} 2024},
	publisher = {Springer Nature Switzerland},
	author = {Xie, Shuzhao and Zhang, Weixiang and Tang, Chen and Bai, Yunpeng and Lu, Rongwei and Ge, Shijia and Wang, Zhi},
	editor = {Leonardis, Aleš and Ricci, Elisa and Roth, Stefan and Russakovsky, Olga and Sattler, Torsten and Varol, Gül},
	year = {2025},
	pages = {434--452},
}

@inproceedings{girish_eagles_2025,
	address = {Cham},
	title = {{EAGLES}: {Efficient} {Accelerated} {3D} {Gaussians} with {Lightweight} {EncodingS}},
	isbn = {978-3-031-73036-8},
	shorttitle = {{EAGLES}},
	doi = {10.1007/978-3-031-73036-8_4},
	language = {en},
	booktitle = {Computer {Vision} – {ECCV} 2024},
	publisher = {Springer Nature Switzerland},
	author = {Girish, Sharath and Gupta, Kamal and Shrivastava, Abhinav},
	editor = {Leonardis, Aleš and Ricci, Elisa and Roth, Stefan and Russakovsky, Olga and Sattler, Torsten and Varol, Gül},
	year = {2025},
	pages = {54--71},
}

@article{bross_overview_2021,
	title = {Overview of the {Versatile} {Video} {Coding} ({VVC}) {Standard} and its {Applications}},
	volume = {31},
	issn = {1051-8215, 1558-2205},
	url = {https://ieeexplore.ieee.org/document/9503377/},
	doi = {10.1109/TCSVT.2021.3101953},
	language = {en},
	number = {10},
	urldate = {2024-02-19},
	journal = {IEEE Transactions on Circuits and Systems for Video Technology},
	author = {Bross, Benjamin and Wang, Ye-Kui and Ye, Yan and Liu, Shan and Chen, Jianle and Sullivan, Gary J. and Ohm, Jens-Rainer},
	month = oct,
	year = {2021},
	pages = {3736--3764},
}

@article{de_queiroz_compression_2016,
	title = {Compression of {3D} {Point} {Clouds} {Using} a {Region}-{Adaptive} {Hierarchical} {Transform}},
	volume = {25},
	issn = {1941-0042},
	url = {https://ieeexplore.ieee.org/abstract/document/7482691},
	doi = {10.1109/TIP.2016.2575005},
	number = {8},
	urldate = {2023-10-13},
	journal = {IEEE Transactions on Image Processing},
	author = {de Queiroz, Ricardo L. and Chou, Philip A.},
	month = aug,
	year = {2016},
	note = {Conference Name: IEEE Transactions on Image Processing},
	pages = {3947--3956},
}

@article{sullivan_overview_2012,
	title = {Overview of the {High} {Efficiency} {Video} {Coding} ({HEVC}) {Standard}},
	volume = {22},
	issn = {1558-2205},
	url = {https://ieeexplore.ieee.org/abstract/document/6316136},
	doi = {10.1109/TCSVT.2012.2221191},
	number = {12},
	urldate = {2024-12-04},
	journal = {IEEE Transactions on Circuits and Systems for Video Technology},
	author = {Sullivan, Gary J. and Ohm, Jens-Rainer and Han, Woo-Jin and Wiegand, Thomas},
	month = dec,
	year = {2012},
	note = {Conference Name: IEEE Transactions on Circuits and Systems for Video Technology},
	pages = {1649--1668},
}

\end{document}